\documentclass[10pt,twocolumn,letterpaper]{article}

\usepackage{times}
\usepackage{epsfig}
\usepackage{graphicx}
\usepackage{amsmath}
\usepackage{amssymb}
\usepackage{subfigure}
\usepackage{url}

\graphicspath{{example_images/}}
\usepackage[pagebackref=true,breaklinks=true,letterpaper=true,colorlinks,bookmarks=false]{hyperref}
\begin{document}

%%%%%%%%% TITLE
\title{Colorization of Natural Images via $L^1$ Optimization}

\author{Alexander Balinsky\\
School of Mathematics \\
Cardiff University, UK\\
{\tt\small BalinskyA@cardiff.ac.uk}
\and
Nassir Mohammad\\
School of Mathematics \\
Cardiff University, UK\\
\& HP Labs, Bristol, UK \\
{\tt\small MohammadN3@cardiff.ac.uk, Nassir.Mohammad@hp.com}
}

\maketitle
% \thispagestyle{empty} % *** Uncomment this line for the final submission

%%%%%%%%% ABSTRACT
\begin{abstract}
Natural images in the colour space $YUV$ have been observed to have a non-Gaussian, heavy tailed distribution (called 'sparse') when the filter 
\begin {equation*}
\gamma(U)(\mathbf r) = U(\mathbf r) - \sum_{ \mathbf s \in N(\mathbf r)} w{(Y)_{\mathbf r \mathbf s}} U(\mathbf s),
\end {equation*}
is applied to the chromacity channel $U$ (and equivalently to $V$), where $w$ is a weighting function constructed from the intensity component $Y$ \cite{Nassir}.  

In this paper we develop Bayesian analysis of the colorization problem using the filter response as a regularization term to arrive at a non-convex optimization problem. This problem is convexified using $L^1$ optimization which often gives the same results for sparse signals \cite{Candes}. It is observed that $L^1$ optimization, in many cases, over-performs the famous colorization algorithm by Levin et al \cite{Levin}.
\end{abstract}

\section{Introduction}
Colorization of natural images has been a long standing problem in image processing. The initial process was invented by Wilson Markle and Brian Hunt and first used in $1970$ to add colour to monochrome footage of the moon from the Apollo mission. One of the main drawbacks of this and subsequent techniques has always been the labour intensity involved and hence the associated costs. This is due to the time-consuming segmentation and tracking of objects in images plus the assignment of colours that all require substantial user involvement. 

However, a recent state of the art method has been proposed by Levin et al in the classical paper \cite{Levin} where colorization is performed by optimization. In the mentioned work a fundamental hypothesis is that areas of similar luminance should have similar colours. This, together with additional colour scribbles placed on the interior regions of objects in the gray image, is used to propagate colour to the rest of the image by minimisation of a quadratic cost function. The result is a visually pleasing image with a reduction in user input. 

A number of recent advancements have since been made to improve the quality and efficiency of the colorization process. These works can roughly be divided into scribble based and example-based colorization. Two highlighted cases are the works presented in \cite{Yatziv} and \cite{Irony}. In the former paper a computationally simple, yet effective, approach is presented which works very fast and can be conveniently used 'on the fly', permitting the user to promptly get the desired results after providing a set of chrominance scribbles. The latter paper develops the method of transferring colour from a segmented example image, and uses the method in \cite{Levin} to produce the finished colorized image. This method has the advantage of not having to rely upon the user's skill or experience in choosing suitable colours for a convincing colorization. 

Partially inspired by the work in \cite{Levin} we study the colorization problem using Bayesian analysis, where we use the response of the filter observed in \cite{Nassir} as a regularization term;

\begin{equation}
\label{Filter}
\gamma(U)(\mathbf r) = U(\bold r) - \sum_{ \mathbf s \in N(\mathbf r)} w{(Y)_{\mathbf r \mathbf s}} U(\mathbf s),
\end {equation}
where $\bold r$ represents a two dimensional point, $N(\bold r)$ a neighborhood (e.g. 3x3 window) of points around $\bold r$, and
$w{(Y)_{\mathbf r \mathbf s}}$ a weighting function. 
The filter is applied to the chromacity channel $U$ (and equivalently to $V$) of individual natural images in the colour space $YUV$. This particular space is chosen as it allows the decoupling of the luminance and colour components of an image. We note here that since the $U$ and $V$ elements are similar, our work is only explained for the $U$ component where analysis of the $V$ component is obtained by substitution. 

The filter response can be modelled by a Generalised Gaussian Distribution (GGD),

\begin{equation}
\label{GGD}
J_{\alpha}(x) = \frac{1}{Z} e^{-|x/s|^\alpha}, 
\end {equation}
where $Z$ is a normalising constant so that the integral of $J_{\alpha}(x)$ is 1, $s$ the scale parameter and $\alpha$ the shape parameter. The GGD gives a Gaussian or Laplacian distribution when $\alpha=2$ or $1$, respectively. When $\alpha<1$ we have a 'heavy tailed' distribution which we call 'sparse'. 

\begin{figure}
\centering	
{\includegraphics[width=8cm, height=5cm]{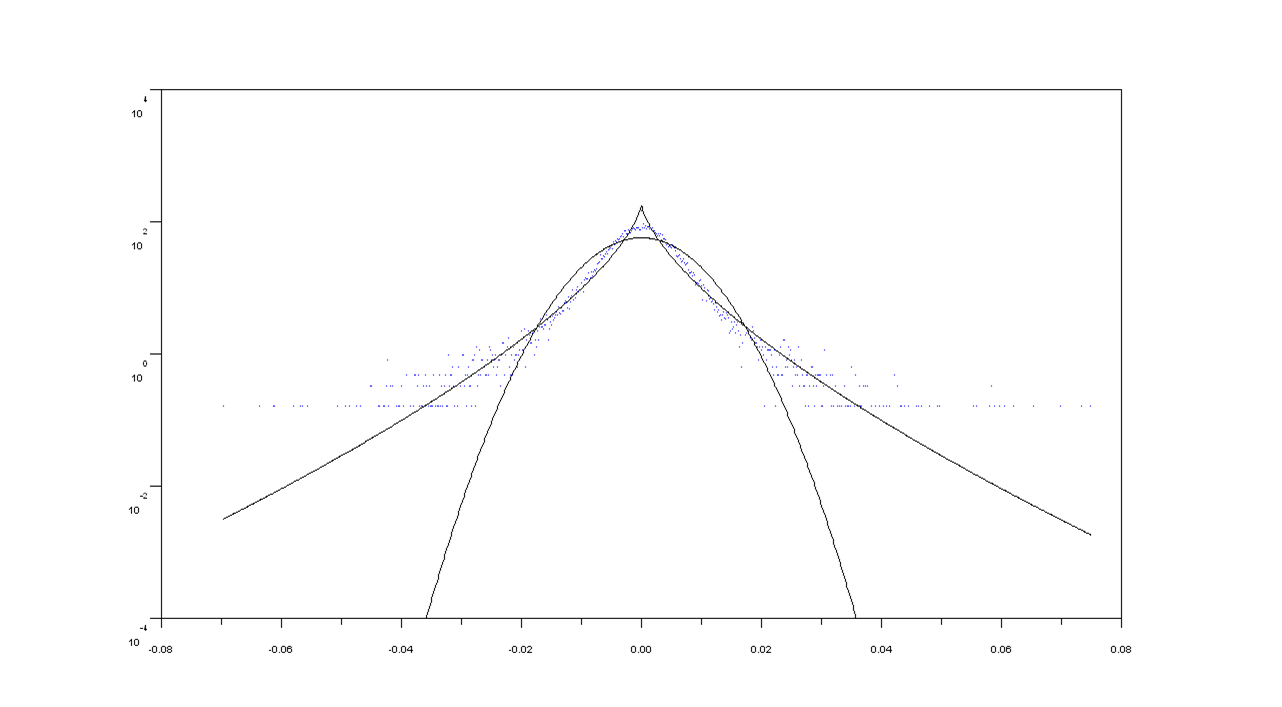}}
\caption{ \footnotesize }
\label{figure4}
\end{figure}

Figure \ref{figure4} shows a typical example (with the vertical axis on a log scale) of the histogram of pixel intensity that is observed after filtering a natural colour image. Fitted to the data is a GGD that takes the form of a sparse distribution function. We also overlay on the figure the classical parabola shaped Gaussian distribution which clearly shows the difference in the tails between the two. Such differences highlight the importance of choosing the correct distribution as a priori knowledge when using Bayesian analysis.

Our key contribution in this paper is tackling the colorization problem using Bayesian analysis with the sparse distribution property of the filter response on natural images acting as a regularization term. We arrive at a non-convex optimisation problem taking the form of a GGD with shape parameter $\alpha<1$. Interestingly, our regularization term indicates that the quadratic cost function (i.e. using $\alpha=2$ in the GGD model) minimised in \cite{Levin} is in fact a Gaussian approximation to the true objective function. 

\section {Bayesian Analysis of the Colorization Problem}
%explanation of bayseian analysis
Our colorization problem is posed as follows: We are given a gray level natural image $Y$ and additional coloured markings $U_o$ on $S$, where $S$ denotes a subset of pixels of the image. We would like find to a $U$ on the whole image s.t. 

\begin{eqnarray*}
&(c1)& U|_s=U_o,\\ 
&(c2)& \mbox{ and the resulting colour image looks natural.}\\
\end{eqnarray*}

Formally we have the following: For any $A$ let us denote by $P_Y(A)$ the conditional probability $P(A|Y)$. Then we wish to maximise 

\begin{equation}
\label{nm1}
\mathbb P_Y(U|U_o).
\end{equation}
Applying Bayes' formula results in maximising

\begin{equation}
\label{nmm2}
\mathbb P_Y(U_o|U)\mathbb P_Y(U),
\end{equation}    
or equivalently to find

\begin{equation}
argmax_U P_Y(U),
\end{equation}
under condition (c1). 

\textbf{Remark:} \textit{If we want to keep $U_o$ exactly then $P_Y(U_0|U)$ is $1$ if $U|_S=U_o$ and $0$ otherwise. If we assume that we can have error in $U_o$ then $P_Y(U_o|U)$ can be modelled as a Gaussian distribution for $U|_S-U_o$.}

We model $P_Y(U)$ by analysing the filter response in \cite{Nassir}. More precisely we use the following expression:

\[
P_Y(U) \propto e^{-\sum |g_i \ast U|^\alpha}
\]
where $g_i$ is the filter operating on the $i$'th pixel in the image. Taking logs leads to an equivalent minimisation objective,

\begin{equation}
\label{min} 
argmin_{U} \ { \sum_i|(g_i \cdot U)|^\alpha} \ \  \mbox{ with } \ {U}|_{S}=U_o.
\end{equation}

With $\alpha=2$, we arrive at the same optimization problem solved by Levin et al in \cite{Levin}. Moreover, this shows that their model gives colorization with Gaussian behavior of the response $g_i \ast U$. However, for all natural images that we have studied, we found that $0<\alpha<2$ but rarely $\alpha>1$ implying the distribution of the filter response to generally be sparse, thus arriving at the correct optimization function. 

\section{Optimization using the $L^1$ norm}

Solving \ref{min} for $\alpha<1$ leads to a non-convex optimization problem that unlike least squares regression has no explicit formula for the solution. We defer the case $\alpha<1$ for future work and instead convexify the problem using $L^1$ optimization which often gives the same results for sparse signals, especially if the size of $|S|$ is small compared to the size of the image \cite{Candes}. Integrating the constraints ${U}|_{S}={U_o}$ into the objective function this problem can be reformulated as an unconstrained one where we seek to minimise, 

\begin{equation}
\label{nm9}
J_1(U) = \sum_i |g_i \cdot U| + \lambda \sum_{i\in S}|U(i) - U_o(i)|,
\end{equation} 
with $i$ running over all image pixels and $S$ the set of pixels that have been marked by colour. This can be constructed into the vectorial form

\begin{equation}
\label{nm10}
||AU-b||_1
\end{equation} 
where $||\cdot||_1$ represents the $L^1$ norm and the $i$'th row of the sparse matrix $A$ corresponds to the filter response of the $i$'th pixel in the image, with $b$ constructed such that \ref{nm9} agrees with \ref{nm10}. 

Using slack variables $\nu_i$ and $\mu_i$, \ref{nm10} can be constructed as a linear program,
\begin{eqnarray}
\label{nm11}
\nonumber & Min \sum_i{\nu_i + \mu_i}, \\
& \mbox{ s.t. }  AU + \nu - \mu = b, \\
& \nonumber \nu_i,\mu_i,U(i),b_i \geq 0
\end{eqnarray} 
The idea here is to find the smallest pairwise addition $\nu_i + \mu_i$, such that their difference is equal to $b(i) - A_{i \to }U$. This occurs  precisely when one of the $\nu_i$ or $\mu_i$ are zero and the other equal to $b(i) - A_{i \to}U$, handling both the positive and negative cases.

\section{Results}

We use the linear programming package 'LIPSOL' \cite{Zhang} available through high-level programming environments, Scilab and Matlab, to solve the problems. Images in the region of $250x250$ pixels take a few minutes to colorize and hence our method is slower than the solvers used in \cite{Levin}. However, our goal here is not to efficiently solve such problems, but only to state the correct optimization problem and to show that when such a problem is solved, the resulting colorized image is of a higher quality.

The results shown in the figures compare the quality of the colorization using the $L^1$ optimisation against the approach of Levin et al. Marking large regions of pixels gives similar results, however, using a much smaller set of marked pixels highlights the difference between the two methods. We note here that since we are only concerned with the correct propagation of colour, and not the choosing of colour, we use the original colour channels of the images for marking colour points on a converted gray image. 

\begin{figure}
\centering	
\subfigure[]{\includegraphics[width=3.9cm, height=3cm]{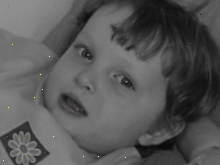}}
\subfigure[]{\includegraphics[width=3.9cm, height=3cm]{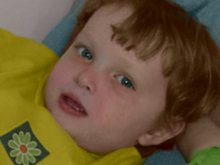}}\\
\subfigure[]{\includegraphics[width=3.9cm, height=3cm]{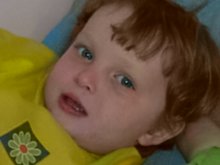}}
\subfigure[]{\includegraphics[width=3.9cm, height=3cm]{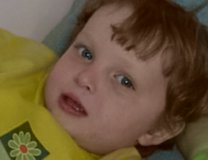}}\\
\subfigure[]{\includegraphics[width=3.9cm, height=3cm]{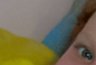}} 
\subfigure[]{\includegraphics[width=3.9cm, height=3cm]{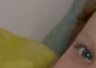}} 

	\caption{ \footnotesize Colorization example. (a) The gray image marked by a sparse set of colour pixels; (b) the original image for reference; (c) colorization using $L^1$ optimization; (d) Levin et al; (e) and (f) closer inspection of (c) and (d) respectively. Here we have an original image from the paper by Levin et al, this time colorized with a much sparser set of pixels placed arbitrarily on the image. We observe more vibrancy in the colours in (c) against the 'washed out' look of the colorization in (d). We also have a sharper result observed upon closer inspection and not an oversmoothed output as usually is the case for assuming a Gaussian prior. }
	
\label{figure1}
\end{figure}

Figure \ref{figure1} shows an example image from the paper by Levin et al. We colorize the same image but using a much sparser set of marked pixels placed arbitrarily in the image regions. (c) shows the improvement in colorization using $L^1$ optimization where we observe a sharper result and not an oversmoothed output as usually is the case for assuming a Gaussian prior. 

\begin{figure}
\centering	
\subfigure[]{\includegraphics[width=3cm, height=4cm]{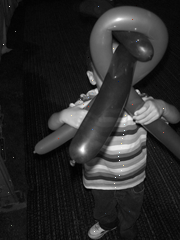}}
\subfigure[]{\includegraphics[width=3cm, height=4cm]{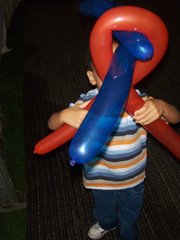}}
\subfigure[]{\includegraphics[width=3cm, height=4cm]{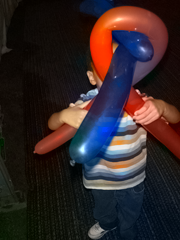}}
\subfigure[]{\includegraphics[width=3cm, height=4cm]{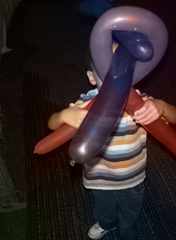}}

	\caption{ \footnotesize Colorization example. Here we have a comparison of the visual quality produced by $L^1$ optimization against the technique used by Levin et al. (a) is an example gray image marked by a sparse set of coloured pixels arbitrarily placed; (b) the original colour image for reference; (c) shows colorization using $L^1$ optimization; (d) method of Levin et al. We observe a more accurate colorization in (c) e.g. the red balloon in the centre of the image is colorized correctly as opposed to the purple colorization in (d). We also observe more vibrant and sharper colours in (c), particularly on the t-shirt.}
\label{figure2}
\end{figure}

Figure \ref{figure2} shows an example of a colorized image. Requiring a relatively few set of marked pixels our method shows good results compared to that of Levin et al's which incorrectly colorizes the red balloon in the centre of the picture as purple. We also observe more vibrancy in the colours in (c) over (d), particularly the stripes on the t-shirt. 

%Figure \ref{figure3} shows an example of a similar output using both methods, however, upon closer inspection we observe sharper results using $L^1$ opimization.

Our results are still a working progress and we are currently improving the efficiency and quality of the colorization. In future we hope to further improve the sharpness and accuracy of the output images whilst significantly reducing the user marked pixels required in the optimization scheme. 

\section {Summary}

The problem of colorizing natural images is tackled using Bayesian analysis where we use the sparse distribution property of the filter response observed in \cite {Nassir} as a regularization term. We arrive at a cost function taking the form of a GGD with shape parameter $\alpha<1$ resulting in a non-convex optimisation problem. Our regularization term indicates that the quadratic cost function (i.e. using $\alpha=2$) minimised in \cite{Levin} is in fact a Gaussian approximation to the true objective function. 

We convexify the non-convex problem using $L^1$ optimization which is often found to give the same results for sparse signals. This problem is formulated as a linear program that can be solved using any standard method; we chose to use the software package 'LIPSOL' available through the programming environments Scilab and Matlab. 

It is observed that $L^1$ optimization, in many cases, over-performs the famous colorization algorithm by Levin et al \cite{Levin}, although at present solving for an equivalent sized image takes longer. However, our aim here is not find the most efficient solvers for such problems, but only to state the correct formulation of the colorization problem. %In future we will solve the non-convex case, which we believe will further improve the quality of results, this will be reported elsewhere.

\section*{Acknowledgments}

We would like to thank Stephen Pollard and Andrew Hunter from HP Labs, Bristol, UK for providing some of the images used in the study and for fruitful discussion. We would also like to thank Anat Levin for the availability of the code used in \cite{Levin} for the purpose of research, and also Yin Zhang for the 'LIPSOL' linear programming solver Matlab package.

This work was supported in part by grants from the Engineering and Physical Sciences Research Council (EPSRC) and Hewlett Packard Labs, awarded through the Smith Institute Knowledge Transfer Network. 

\newpage

\end{document}